\documentclass[10pt,journal]{IEEEtran}

\usepackage{cite}
\usepackage{amsmath,amssymb,amsfonts}
\usepackage{algorithmic}
\usepackage{graphicx}
\usepackage{textcomp}
\usepackage{multirow}
\usepackage{adjustbox}
\usepackage{longtable}
\usepackage{makecell}
\usepackage{booktabs}
\usepackage{ltablex}
\usepackage{tabularx}
\usepackage{array}
\usepackage{caption}

\keepXColumns
\def\BibTeX{{\rm B\kern-.05em{\sc i\kern-.025em b}\kern-.08em
    T\kern-.1667em\lower.7ex\hbox{E}\kern-.125emX}}

\begin{document}

\title{MCL: Meta Convolution Layer}

\author{Naim Reza, Md Al Amin, and Ho Yub Jung%
\thanks{Naim Reza, Md Al Amin, and Ho Yub Jung are with the Department of Computer Engineering, Chosun University, Gwangju 61452, Republic of Korea (e-mail: naim@chosun.kr; ahmedalamin@chosun.ac.kr; hoyub@chosun.ac.kr).}%
\thanks{Corresponding author: Ho Yub Jung (e-mail: hoyub@chosun.ac.kr).}}

\maketitle

\begin{abstract}

Dynamic convolution enhances convolutional neural networks (CNNs) by adapting kernels to input content, but it expresses the effective kernel as a linear mixture of a small number of basis kernels, which limits expressivity and complicates optimization as the mixture size grows. In this work, we revisit dynamic convolution from a functional perspective and propose the \emph{Meta Convolution Layer} (MCL), which directly models the convolutional kernel as an input-conditioned function $\mathbf{W}(x)$ realized via a high-order polynomial expansion. Leveraging nested residual blocks inspired by deep polynomial networks, MCL implements a structured polynomial meta-network that generates a single input-adaptive kernel, thereby decoupling representational power from the explicit number of mixture kernels and alleviating training instability. MCL is a plug-in addition with standard convolutions and can be seamlessly integrated into both CNN and transformer backbones. 
Experimental evaluation shows that adding MCL improves the Top-1 accuracy of Resnet-18, Resnet-50 and ResNet-101 by $6.61\%$, $3.42\%$ and $3.05\%$ on the ImageNet dataset. Moreover, the proposed method significantly boosts the accuracy of Resnet and Wide-Resnet variants on CIFAR-10 and CIFAR-100 datasets. Additionally, the proposed method outperforms previous methods on fine-grained visual classification tasks using Swin and ViT backbones.
These results demonstrate that high-order polynomial kernel generation is a powerful and scalable alternative to linear mixture based dynamic convolution.
 
\end{abstract}


\section{Introduction}
\label{sec:intro}

Deep convolutional neural networks (CNNs) have achieved remarkable success across a wide range of visual recognition tasks, including large-scale image classification, object detection and fine-grained recognition\cite{alexnet, mobilenet, fgvc}. This success is largely attributed to the representational power of convolutional layers, which learn hierarchical feature representations by repeatedly applying spatially local, channel-wise shared kernels to input feature maps. In the standard design, each convolutional layer uses a single static kernel that is shared across all input samples, and the expressiveness of the model is mainly increased by stacking more layers or widening the network. While effective, such static kernels cannot adapt to the diverse and input-dependent patterns present in natural images.

To enhance the adaptability of CNNs, a large body of work has focused on feature recalibration through attention mechanisms. Channel attention and spatial attention modules, such as squeeze-and-excitation (SE) blocks\cite{se}, convolutional block attention modules (CBAM)\cite{cbam}, efficient channel attention (ECA)\cite{eca}, and related variants, learn input-dependent importance weights to re-scale feature channels or spatial locations, thereby emphasizing informative responses while suppressing less relevant ones. These mechanisms significantly improve the performance of baseline CNNs with modest parameter overhead. However, they operate on feature activations while keeping the convolutional kernels themselves fixed, and thus cannot fully exploit the potential of input-adaptive kernel generation.

Dynamic convolution takes a more direct approach by learning input-dependent kernels. Instead of using a single static kernel per layer, dynamic convolution approximates the effective kernel as a linear mixture of $n$ candidate kernels, each modulated by an attention coefficient predicted from the input\cite{dynamic_conv, dcd}. This design enables the network to instantiate different effective kernels for different inputs, improving the representation capacity especially for lightweight models. Subsequent works generalize this idea along different dimensions of the kernel space. For example, CondConv\cite{condconv} and DY-Conv\cite{dynamic_conv} learn input-conditioned mixtures of multiple kernels, while Omni-dimensional Dynamic Convolution (ODConv)\cite{odconv} extends the dynamic mechanism to simultaneously modulate kernel number, spatial size, input channel and output channel dimensions. KernelWarehouse\cite{kw} further rethinks dynamic convolution by partitioning kernels into a shared ``warehouse'' and assembling them with learnable attention, enabling parameter sharing across layers and larger effective kernel sets under a fixed parameter budget.

Despite these advances, existing dynamic convolution methods share a common limitation, the effective kernel is still expressed as a \emph{linear} combination of a finite set of basis kernels. In practice, the number of kernels $n$ is often constrained to a small value (typically $n < 10$) to control parameter growth and computational cost. Increasing $n$ improves expressivity but it also scales parameters and FLOPs linearly, and makes optimization difficult, often requiring careful temperature scheduling for the attention softmax and specialized initialization schemes to stabilize training \cite{dynamic_conv, dcd, kw}. Even with advanced designs such as KernelWarehouse\cite{kw} or ODConv\cite{odconv}, the underlying combination rule remains linear, which may limit the ability to capture high-order, input-dependent interactions in large and wide networks.

In parallel, polynomial neural networks\cite{pinet} and related functional approximators\cite{reg-poly} have shown that high-order polynomial expansions can provide powerful and compact representations when implemented with appropriate structure. Deep Polynomial Networks ($\Pi$-Nets)\cite{pinet} introduce nested coupled polynomial (NCP) and NCP-Skip decompositions to realize $2^N$-th order polynomial expansions using only $N$ residual-like blocks, providing a principled way to model rich interactions with controlled parameter growth. These works suggest that structured polynomial expansions, when embedded in residual architectures, can approximate complex functions efficiently and remain amenable to gradient-based optimization.

Motivated by these studies, we revisit dynamic convolution from a functional perspective and consider the convolutional kernel as an explicit input-conditioned function $\mathbf{W}(x)$ instead of a linear mixture of a few static kernels. We propose \emph{Meta Convolution Layer} (MCL), which models $\mathbf{W}(x)$ via a high-order polynomial expansion implemented through nested residual blocks following the NCP-Skip decomposition. Instead of aggregating $n$ kernels with attention, MCL uses a polynomial meta-network to directly generate a single input-adaptive kernel that captures high-order interactions in the feature space. This design mitigates the difficulty of jointly optimizing many kernels, decouples expressivity from the explicit kernel count $n$, and naturally scales to deeper and wider backbones.

At a high level, MCL bridges dynamic convolution and deep polynomial networks. It preserves the plug-and-play nature of dynamic convolution modules while replacing linear mixtures with structured polynomial expansions of the kernel. We validate MCL across multiple regimes, including CIFAR\cite{cifar}, ImageNet\cite{imagenet}, and fine-grained visual classification (FGVC) benchmarks with both CNN and transformer backbones, demonstrating consistent improvements over strong baselines and existing dynamic convolution methods.

The main contributions of this work are summarized as follows:
\begin{itemize}
    \item We propose the Meta Convolution Layer (MCL), which directly estimates the convolution kernel via a high-order polynomial expansion of input features and generates a single input-adaptive kernel.
    \item We show that the proposed MCL with a polynomial adaptive kernel is easy to train without hyperparameter tuning. This is the main advantage over the previous attention-based linear kernel mixtures where softmax temperature ($\tau$) scheduling is required for stable optimization. 
    \item We demonstrate that MCL can be seamlessly integrated into standard CNN backbones and transformer architectures, consistently improving performance on CIFAR, ImageNet, and multiple FGVC benchmarks, while offering favorable accuracy--parameter trade-offs compared to prior dynamic and attention-based modules.
    \item We provide empirical analyses on polynomial order via kernel size and meta-block depth, parameter efficiency, and computational cost, highlighting the scalability and practical applicability of MCL as a general plug-in module alongside static convolutions.
\end{itemize}

\section{Related Works}
\label{sec:related_works}

\paragraph{Dynamic Convolution and Attention over Kernels.}

Dynamic convolution generalizes standard convolution by making the kernel input-dependent. Early approaches such as Dynamic Convolution\cite{dynamic_conv} and CondConv\cite{condconv} construct the effective kernel as a linear combination of multiple candidate kernels, where the mixture weights are produced by a lightweight attention function conditioned on the input feature map. DY-Conv\cite{dcd} further refines this paradigm by adopting more flexible attention formulations and regularization strategies to improve training stability. These designs have been particularly effective for compact architectures such as MobileNetV2\cite{mobilenet,mobilenetv2,mobilenetv3} and shallow ResNet\cite{resnet} variants, achieving improved accuracy with moderate increases in parameters and FLOPs.

Subsequent research extends kernel attention to richer factorized spaces. ODConv\cite{odconv} introduces omni-dimensional attentions along kernel number, spatial size, input channels, and output channels, enabling more expressive modulation and yielding consistent improvements on large-scale benchmarks such as ImageNet\cite{imagenet} and MS-COCO\cite{coco}. Complementary analyses revisit dynamic convolution through matrix decomposition to study low-rank structure, parameter efficiency, and regularization effects\cite{dcd}. KernelWarehouse (KW)\cite{kw} further decouples kernel construction from kernel usage by organizing kernels into a shared warehouse of kernel cells and assembling layer-specific kernels via input-dependent scalar attentions, allowing the effective mixture size to scale beyond the typical $n<10$ regime under a fixed parameter budget. Nevertheless, KW and related methods still rely on linear mixtures of basis elements and often require careful initialization and temperature scheduling to maintain stable attention distributions during optimization.

\paragraph{Feature Recalibration and Attention in CNNs.}

Feature recalibration modules improve CNNs by modulating activations while keeping kernels fixed. Squeeze-and-Excitation (SE)\cite{se} introduces channel-wise attention via global pooling and a lightweight bottleneck to re-scale channels, while CBAM\cite{cbam} augments this with sequential channel and spatial attention. ECA\cite{eca} simplifies channel attention using local cross-channel interactions without dimensionality reduction. Related variants such as CGC and WeightNet explore alternative parameterizations and normalization strategies for attention over channels or weights\cite{cgc, weightnet}. Although these modules are complementary to dynamic convolution, they do not explicitly generate input-conditioned kernels. In contrast, our approach targets kernel generation itself: rather than enriching feature responses with fixed kernels, we directly increase kernel expressivity via an explicit functional form.

\paragraph{Polynomial Neural Networks and Dynamic Parameterization.}

Polynomial neural networks offer a principled mechanism to capture high-order interactions. Deep Polynomial Networks ($\Pi$-Nets) address the combinatorial growth of classical polynomial models by using nested coupled polynomial (NCP) and NCP-Skip decompositions, realizing $2^N$-th order expansions with only $N$ residual-like modules\cite{pinet}. Building on these insights, we design MCL to generate convolutional kernels through NCP-Skip style recursions in the \emph{kernel space}, where the meta-network maps the current feature map to the kernel weights $\mathbf{W}(x)$. This contrasts with standard dynamic convolution, which constrains $\mathbf{W}(x)$ to linear mixtures of a few kernels, and relates to broader dynamic parameterization schemes such as HyperNetworks and one-shot weight generation\cite{hypernet, bertinetto}. 

Beyond generating a single kernel, some other work goes even further. These works represent things like images or scenes as continuous functions, instead of storing them as fixed grids of values. Implicit neural representations (INRs) do this by treating images, videos, or 3D scenes as functions that map coordinates to values, as done in Neural Radiance Fields~\cite{mildenhall2021nerf} and SIREN~\cite{sitzmann2020implicit}. Recent work has also applied this idea to image classification, closing some of the gap with regular pixel-based methods~\cite{gielisse2025end}. INRs and MCL are similar in one way. Both replace a fixed, explicit representation with a function that is evaluated only when needed. But they differ in what that function takes as input. INRs learn a function of spatial coordinates, built for one specific scene. MCL instead learns a function of the input feature map itself, and this function produces a single kernel that is shared across every spatial position in that input. HyperNetwork research has also grown beyond the original per-layer embedding approach of Ha et al.~\cite{hypernet}. Some newer work uses graph-based hypernetworks that can generate weights for many different target architectures~\cite{knyazev2021parameter}. Other work uses transformer-based hypernetworks that condition on a small support set of examples for few-shot learning~\cite{zhmoginov2022hypertransformer}. More recently, researchers have shown that pretrained foundation models can further improve how transformer-based hypernetworks generate weights~\cite{gu2025foundation}. These approaches share MCL's goal of avoiding a large explicit weight tensor, but typically generate weights for entire layers or networks from a compact conditioning signal, whereas MCL restricts weight generation to a single convolutional kernel conditioned directly on the local input feature map.

\paragraph{Transformers and Fine-Grained Visual Recognition.}

Transformer backbones such as ViT and Swin become dominant for fine-grained recognition\cite{vit, swin}, recent FGVC methods incorporate specialized token or part-based mechanisms\cite{acc-vit, ielt, mpsa}. Our experiments integrate MCL into ViT and Swin for FGVC datasets and compare against representative approaches (e.g., IELT, MP-FGVC, ACC-ViT, MPSA)\cite{ielt, mp-fgvc, acc-vit, mpsa}, demonstrating that polynomially generated kernels provide consistent gains beyond purely convolutional settings.

Adaptive mechanisms are also central to Vision Transformers themselves. Self-attention generates input-dependent mixing weights over tokens through a softmax, which, similar to dynamic convolution, expresses its output as a linear combination of existing elements rather than as a directly generated function of the input. Other adaptive components used in recent ViT and Swin variants, such as dynamic position bias and lightweight adapter modules, similarly adjust existing weights or biases based on the input rather than generating a new kernel directly. MCL differs from these mechanisms in that it produces an explicit input-dependent convolutional kernel through a polynomial expansion, rather than reweighting a fixed set of tokens or parameters. In our experiments, MCL is inserted as an additional module alongside standard self-attention layers in Swin-B and ViT-B/16, indicating that polynomial kernel generation is complementary to, rather than a replacement for, transformer-native adaptive mechanisms.

\section{Methodology}
\label{sec:method}

In this section, we will outline dynamic convolution and then describe the proposed meta convolution method in detail. 

\subsection{Motivation}
\label{subsec:motivation}

For a convolution layer, let $x \in \mathbb{R}^{c\times h\times w}$ and $y \in \mathbb{R}^{f\times h\times w}$ having $c$ feature channels and $f$ output channels respectively, where $h\times w$ denotes the feature size. A traditional convolution, $y = \mathbf{W}\otimes x$, uses a static kernel $\mathbf{W} \in \mathbb{R}^{f\times c\times k\times k}$ consisting of $f$ filters of spatial size $k\times k$. Dynamic convolution \cite{dynamic_conv, dcd} approximates this $\mathbf{W}$ using aggregation of a linear mixture of $n$ number of same sized kernels, $\mathbf{W_1}, \dots,\mathbf{W_n}$, weighted by $n$ number of input-dependent attention scores, $\alpha_1,\dots,\alpha_n$, generated by an attention module $\phi(x)$. Thus, dynamic convolution is formulated as follows:
\begin{equation}
    y = \sum_{i=1}^{n} \alpha_i \mathbf{W}_i \otimes x + \sum_{i=1}^{n} \alpha_i \mathbf{b}_i
    \label{eq:dy-conv}
\end{equation}
where, $\mathbf{b}$ is the bias vector and $\mathbf{b} \in \mathbb{R}^f$. Although, this non-linear representation of the kernel can learn richer representation but it requires a large number of kernels to achieve this. 
Moreover, jointly optimizing many kernels is difficult and creates training instability by requiring near-uniform attention at the initial stage of training, which is facilitated by using a softmax annealing temperature $\tau$ as $\alpha_i = \frac{exp(z_i/\tau)}{\sum_j exp(z_j/\tau)}$, where, $z_i$ is the output of the final FC layer of the attention module $\phi(x)$.


These shortcomings motivate us to utilize a polynomial expansion to estimate a single kernel to learn high-dimensional representation rather than using multiple kernels. Two practical factors further distinguish MCL from prior dynamic convolution methods. First, MCL incorporates a normalizing scaling factor $\gamma$ in Eq.~\eqref{eq:meta_conv}, which ensures stable variance at initialization and is absent in prior linear mixture approaches. Second, MCL omits softmax-weighted kernel aggregation, thereby avoiding the large zero-gradient regions that typically hinder optimization in both dynamic convolution and attention-based architectures.


\begin{figure*}
    \centering
    \includegraphics[width=\linewidth]{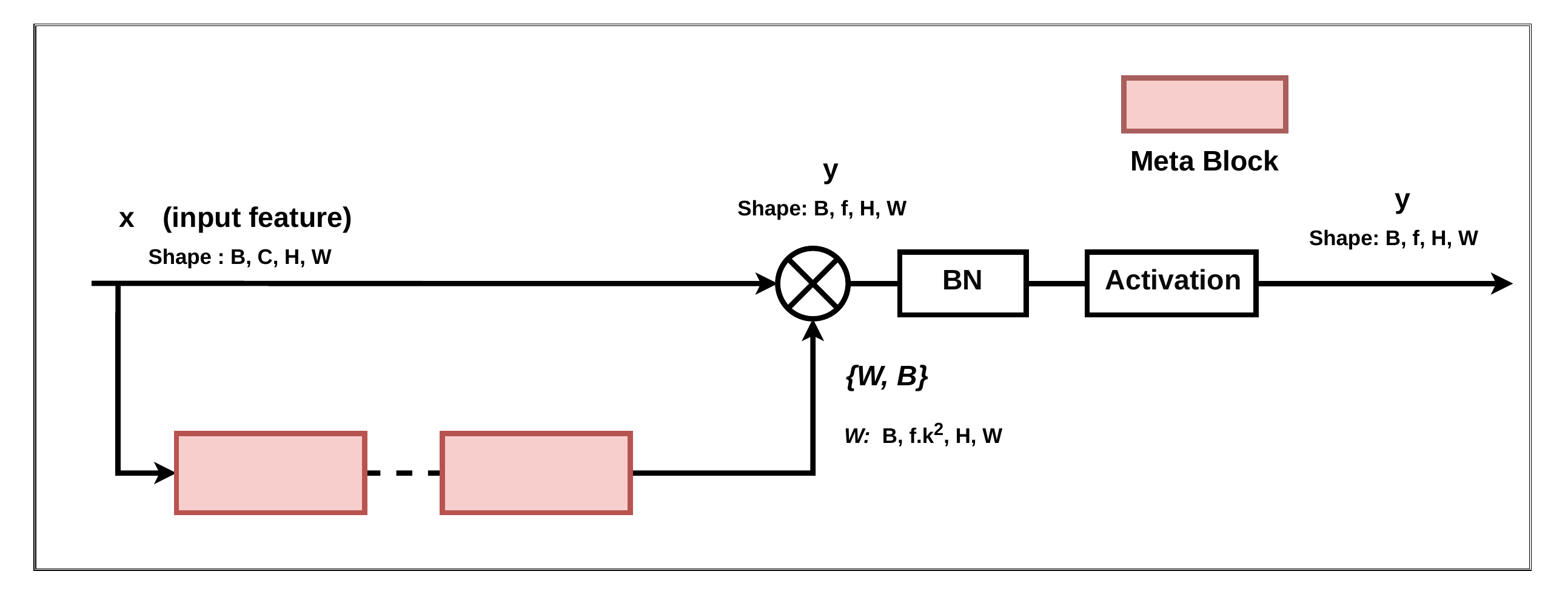}
    \caption{Schematic diagram of the proposed MCL method.}
    \label{fig:method}
\end{figure*}

\subsection{Meta kernel via Polynomial expansion}
\label{subsec:meta-kernel}

To address the limitations of dynamic convolution in jointly optimizing multiple kernels, we propose to estimate the convolution kernel $\mathbf{W}$ directly via a high-order polynomial expansion from input $x$, rather than as a weighted mixture of kernels. Specifically, we replace the linear mixture formulation of Eq.~\ref{eq:dy-conv}, with a formulation where $\mathbf{W}$ is a function of the input $x$, parameterized via a structured polynomial expansion. 

Inspired by the Nested Coupled Polynomial (NCP) and NCP-Skip decomposition introduced in the $\Pi$-Net framework~\cite{pinet}, we define $\mathbf{W}(x)$ recursively through a set of latent features $\{x_n\}_{n=1}^{N}$ encoding polynomial interactions up to order $N$. A sequence of recursive hidden states $x_n \in \mathbb{R}^{k}$ can be defined as:
\begin{align}
x_1 &= (A_1^T x) \odot (B_1^T b_1) \label{eq:poly_x1} \\
x_n &= (A_n^T x) \odot (S_n^T x_{n-1} + B_n^T b_n) + x_{n-1}; \scalebox{0.8}{\text{for } n = 2, \dots, N} \label{eq:poly_xn}
\end{align} where, $A_n \in \mathbb{R}^{c \times k}$ projects the input to a latent space, $S_n \in \mathbb{R}^{k \times k}$ captures recursive dependencies across orders, $B_n \in \mathbb{R}^{\omega \times k}$ and $b_n \in \mathbb{R}^{\omega}$ define the bias transformations and $\odot$ denotes the Hadamard product. The matrix multiplication can represent various linear operations including convolutions. Note that all the components of Eq.~\eqref{eq:poly_x1} and Eq.~\eqref{eq:poly_xn} are differentiable and can be implemented within a residual block. Eq.~\eqref{eq:poly_xn} suggests that the output of a residual block would be a quadratic expansion of the polynomial. Stacking $N$ such polynomials results in an overall $2^N$ order of expansion. Therefore, we can simply stack $N$ number of residual block sequentially to achieve our desired order of expansion of the polynomial. We refer the reader to \cite{pinet} for more insights on this subject matter. 

\begin{figure*}
    \centering
    \includegraphics[width=\linewidth]{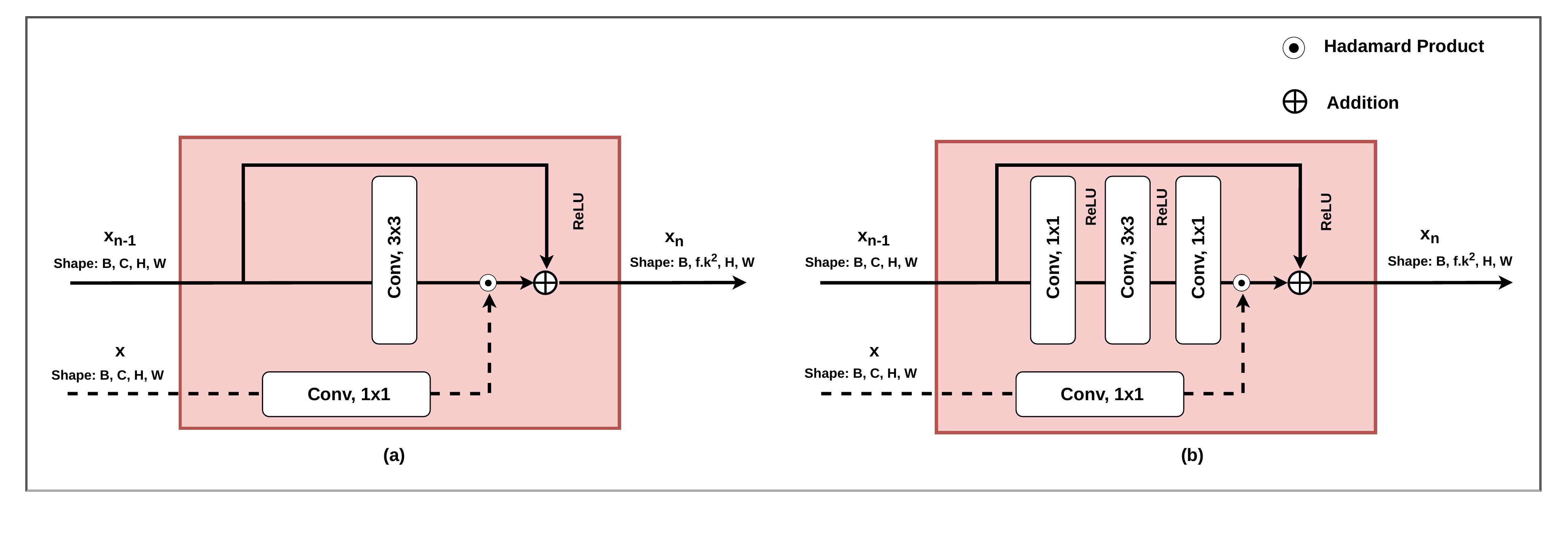}
    \caption{Schematic Diagram of (a) Basic Meta-Block and (b) Compact Meta-Block. The bias term is omitted for clarity.}
    \label{fig:meta-block}
\end{figure*}

Let $x \in \mathbb{R}^{c \times h \times w}$ denote the input feature map and $\mathbf{W}(x) \in \mathbb{R}^{f \times c \times k \times k}$ the generated convolution kernel. Eq.~\eqref{eq:poly_x1}--\eqref{eq:poly_xn} define the polynomial recursion through linear projections, Hadamard products, and residual addition. In practice, we instantiate each linear map (e.g., $A_n^\top(\cdot)$ and $S_n^\top(\cdot)$) with a standard Conv--BN--ReLU operator, as shown in Fig.~\ref{fig:meta-block}. Concretely, for order $n$, we implement the projections in Eq.~\eqref{eq:poly_x1}--\eqref{eq:poly_xn} as,
\begin{align}
\tilde{A}_n^\top(x) &\triangleq \sigma(\mathrm{BN}(\mathrm{Conv}_{A_n}(x))), \label{eq:proj_a}\\
\tilde{S}_n^\top(x_{n-1}) &\triangleq \sigma(\mathrm{BN}(\mathrm{Conv}_{S_n}(x_{n-1}))), \label{eq:proj_s}\\
\tilde{B}_n^\top(b_n) &\triangleq \sigma(\mathrm{BN}(\mathrm{Conv}_{B_n}(b_n))) \label{eq:proj_b}
\end{align}
where $\sigma(\cdot)$ denotes ReLU. Substituting these into Eq.~\eqref{eq:poly_x1}--\eqref{eq:poly_xn} yields the implementation-level recursion
\begin{align}
x_1 &= \tilde{A}_1^\top(x)\ \odot\ \tilde{B}_1^\top(b_1), \label{eq:proj_x1}\\
x_n &= \tilde{A}_n^\top(x)\ \odot\ \big(\tilde{S}_n^\top(x_{n-1}) + \tilde{B}_n^\top(b_n)\big) + x_{n-1} \label{eq:proj_xn}
\end{align}
$for\quad n=2 \ldots N,$ this formulation of the meta-blocks realizes the same multiplicative and residual topology as Eq.~\eqref{eq:poly_x1}--\eqref{eq:poly_xn}, while each ``matrix'' operator is implemented by a small convolutional sub-network. Importantly, if $\sigma$ is replaced by the identity, the implementation reduces exactly to the pure polynomial form in Eq.~\eqref{eq:poly_x1}--\eqref{eq:poly_xn}.

After stacking $N$ meta-blocks, the final state $x_N$ is linearly projected and reshaped to realize the final meta-polynomial kernel,

\begin{equation}
\mathbf{W}(x) = \text{reshape}(C x_N + \beta)
\label{eq:poly_w}
\end{equation}

where, $C$ and $\beta$ are linear transformations. This formulation estimates $\mathbf{W}$ as a single input-adaptive kernel capturing high-order interactions via nested polynomial transformations, providing an efficient and expressive alternative to traditional dynamic convolutions. With ReLU activation retained in Eq.~\eqref{eq:proj_a}--\eqref{eq:proj_b}, $\mathbf{W}(x)$ becomes a \emph{piecewise-polynomial} function that makes polynomial interactions gated by piecewise-linear activations. This hybrid design is consistent with prior polynomial-network practice that explicitly study settings both with and without activations, and report that keeping standard nonlinearities can mitigate overfitting and stabilize training~\cite{pinet, reg-poly}. All projection operators, $\{A_n,S_n,B_n,b_n\}_{n=1}^N$ that parameterize the meta-polynomial recursion, and $C,\beta$ that parameterize the final kernel projection are learnable.

\subsection{Meta Convolution Layer}
\label{subsec:mcl}

Given the input feature $x$ and its associated meta–polynomial kernel $\mathbf{W}(x)$ from Eq.~\eqref{eq:poly_w}, we compute a normalizing scaling factor $\gamma$ as follows, analogous to Xavier and He-initialization~\cite{xaviar, he}, to ensure equal input and output variance at initialization:
\begin{equation}
    \gamma = \sqrt{\frac{2}{c\times k^2 + f}}
\end{equation}


where, $c$, $k$, and $f$ denote the input channel size, meta-kernel size, and output channel size, respectively. Note that $\gamma$ is a fixed architectural constant fully determined by these dimensions. Omitting this normalization can cause training instability, as the unnormalized meta-kernel outputs would have variance that scales with the network dimensions.

\begin{equation}
y = \gamma\mathbf{W}(x) \otimes x + \mathbf{B}
\label{eq:meta_conv}
\end{equation}

where $\mathbf{B} \in \mathbb{R}^{f}$ is a learnable bias, and $\otimes$ denotes convolution. Following this, we use batch normalization and activation function like traditional convolution as shown in Fig.~\ref{fig:method}. This formulation is consistent with the dynamic convolution notation in Eq.~\eqref{eq:dy-conv}, but with \emph{a single} input-dependent kernel $\mathbf{W}(x)$ generated by a polynomial meta-network instead of a mixture of multiple static kernels. The learnable parameters are optimized end-to-end with the backbone network and thus easy to optimize.

\subsection{Optimization Behavior and Stability}
Here, we discuss why MCL remains stable to optimize despite generating a high-order polynomial kernel.

The scaling factor $\gamma$ in Eq. (10) serves the same purpose as Xavier and He initialization for standard convolution. Without $\gamma$, the variance of $\mathbf{W}(x)$ grows with the product of input channels, kernel size, and output channels, so the gradient magnitude at the convolution in Eq. (11) would scale with these dimensions and destabilize training at initialization, particularly for larger meta-kernel sizes $k$. Fixing $\gamma$ as in Eq. (10) removes this dependency and keeps the expected output variance constant across layer widths. Each meta-block in Eq. (7)-(8) also adds the previous hidden state back onto its transformed output, the same way a residual block does. Because of this, the gradient going back to each hidden state always has a direct, unblocked path, no matter how the multiplicative Hadamard term inside the block behaves. This is why the polynomial order can grow as $2^N$ while the number of steps only grows linearly with $N$. Stacking more meta-blocks does not make the vanishing or exploding gradient problem worse, unlike typical deep multiplicative chains.

Finally, the attention mixture in Eq. (1) used by prior dynamic convolution methods needs a softmax over the kernel logits. When this softmax becomes close to one-hot, the gradient for the unused kernels drops to nearly zero, so those kernels stop learning. MCL does not use softmax over kernels, so this problem does not happen. Instead, the ReLU activations in Eq. (4)-(6) pass each polynomial term through in a piecewise-linear way, so a gradient is either passed through fully or blocked, not slowly reduced to zero like with softmax. This matches the stable training we see in the standard deviations reported in Table \ref{tab:cifar} and the convergence curves in the appendix.

\subsection{Implementation}
\label{subsec:implementation}

Unlike prior dynamic convolution approaches that replace the $3\times 3$ operators throughout the backbone, we integrate only one MCL at the end of the final feature-extraction stage. Since stacking $N$ meta-blocks within a single MCL results in a $2^{N}$-order polynomial expansion, a single MCL suffices to model high-order interactions while keeping the backbone unchanged. The output of MCL is then forwarded to the final fully connected classifier layer for prediction. The plug-and-play property of MCL means that it does not itself introduce training instability when attached to any backbone. MCL is not designed to resolve pre-existing optimization fragility in the host architecture, such as the difficulty of training ViTs from scratch on small datasets.

To instantiate each meta-block, we adopt the residual-style configuration in Fig.~\ref{fig:meta-block}(a) and denote the MCL kernel size by $k$ and the polynomial expansion order by $n$. While larger $k$ and higher-order expansions ($n>1$) improve accuracy, they also increase parameter cost as later shown in appendix~\ref{sec:appendix-k-n}. To control this overhead, we employ the compact meta-block in Fig.~\ref{fig:meta-block}(b), where the expensive $3\times 3$ operation is replaced with $1\times 1$ convolutions. In all experiments, we set $n=2$ (two meta-blocks), yielding a $4^{\text{th}}$-order expansion, and use symmetric meta-kernels (e.g. $3\times 3$ and $5\times 5$). To preserve the same spatial resolution of the backbone feature at the output of MCL, we fix stride to $1$ and set padding to $0/1/2$ for $k=1/3/5$, respectively.

\section{Experiments}
\label{sec:experiments}

We evaluate the proposed Meta Convolution Layer (MCL) on CIFAR~\cite{cifar}, ImageNet-1K~\cite{imagenet}, and four fine-grained visual classification (FGVC) benchmarks to assess accuracy gains, scalability, and transferability.

\subsection{Image Classification on CIFAR Dataset}
\label{subsec:cifar}


\begin{table*}[t]
\centering
\caption{Classification accuracy (\%) on CIFAR-10 and CIFAR-100 using ResNet and
WideResNet backbones with and without the proposed MCL\@. We report Top-1 accuracy
$\pm$ STD of 3 independent runs for different meta-kernel sizes $k$ and training
settings.}
\label{tab:cifar}
\resizebox{\linewidth}{!}{%
\begin{tabular}{l|ccc|cccc|ccc|cccc}
\toprule
\multirow{3}{*}{\textbf{Model}}
  & \multicolumn{7}{c|}{\textbf{CIFAR10}}
  & \multicolumn{7}{c}{\textbf{CIFAR100}} \\
\cmidrule(lr){2-8}\cmidrule(lr){9-15}
  & \multirow{2}{*}{Baseline}
  & \multirow{2}{*}{Baseline + MCL}
  & \multirow{2}{*}{Baseline + Aug}
  & \multicolumn{4}{c|}{Baseline + Aug + MCL}
  & \multirow{2}{*}{Baseline}
  & \multirow{2}{*}{Baseline + MCL}
  & \multirow{2}{*}{Baseline + Aug}
  & \multicolumn{4}{c}{Baseline + Aug + MCL} \\
\cmidrule(lr){5-8}\cmidrule(lr){12-15}
  & & &
  & $k{=}1$ & $k{=}3$ & $k{=}3(b)$ & $k{=}5(b)$
  & & &
  & $k{=}1$ & $k{=}3$ & $k{=}3(b)$ & $k{=}5(b)$ \\
\midrule
ResNet20
  & 91.43
  & $93.52\pm0.07$
  & 92.33
  & $94.53\pm0.18$
  & $\mathbf{95.32\pm0.08}$
  & $94.66\pm0.09$
  & $95.20\pm0.05$
  & 69.95
  & $70.41\pm0.05$
  & 68.89
  & $72.83\pm0.19$
  & $77.97\pm0.23$
  & $77.32\pm0.27$
  & $\mathbf{78.58\pm0.23}$ \\
ResNet56
  & 93.03
  & $94.32\pm0.08$
  & 94.31
  & $95.81\pm0.13$
  & $96.09\pm0.12$
  & $95.77\pm0.04$
  & $\mathbf{96.12\pm0.08}$
  & 71.97
  & $73.54\pm0.24$
  & 75.26
  & $78.26\pm0.16$
  & $80.37\pm0.08$
  & $79.66\pm0.18$
  & $\mathbf{81.14\pm0.13}$ \\
ResNet110
  & 93.57
  & $94.72\pm0.18$
  & 94.90
  & $96.16\pm0.00$
  & $\mathbf{96.57\pm0.06}$
  & $96.32\pm0.07$
  & $96.43\pm0.15$
  & 74.80
  & $75.00\pm0.18$
  & 78.11
  & $79.73\pm0.14$
  & $81.81\pm0.19$
  & $80.93\pm0.11$
  & $\mathbf{82.25\pm0.05}$ \\
WideResNet16-10
  & 95.44
  & $96.01\pm0.04$
  & 96.80
  & $97.71\pm0.00$
  & $97.78\pm0.04$
  & $\mathbf{97.80\pm0.07}$
  & $97.65\pm0.03$
  & 78.41
  & $79.96\pm0.26$
  & 82.86
  & $84.62\pm0.08$
  & $85.10\pm0.25$
  & $\mathbf{85.62\pm0.11}$
  & $85.44\pm0.04$ \\
WideResNet28-10
  & 96.00
  & $96.25\pm0.02$
  & 97.41
  & $\mathbf{98.05\pm0.11}$
  & $97.93\pm0.01$
  & $97.98\pm0.03$
  & $97.98\pm0.10$
  & 80.75
  & $80.87\pm0.23$
  & 83.50
  & $85.26\pm0.09$
  & $86.00\pm0.13$
  & $85.90\pm0.03$
  & $\mathbf{86.20\pm0.09}$ \\
\bottomrule
\end{tabular}%
}
\end{table*}

\begin{figure}
    \centering
    \includegraphics[width=0.8\linewidth]{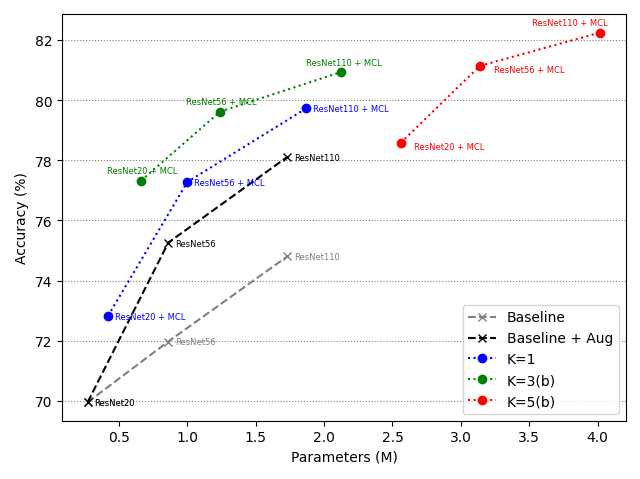}
    \caption{This figure illustrates the comparison between accuracy vs parameters (M) for varying meta-kernel sizes of MCL on the CIFAR-100 dataset.}
    \label{fig:c100-accvsparam}
\end{figure}

We study MCL on CIFAR-10/100~\cite{cifar} using ResNet-20/56/110~\cite{resnet} and WideResNet-16-10/28-10~\cite{wrn}. All models are trained for 200 epochs with SGD (momentum 0.9, weight decay $5\times10^{-4}$), batch size 128, and a cosine learning-rate schedule from 0.1. We apply standard CIFAR augmentation, random crop with 4-pixel padding, horizontal flip, normalization, and additionally report results with MixUp~\cite{cutmix} as stronger data augmentation. Unless stated otherwise, we set the polynomial order to $n=2$ and vary the meta-kernel size $k$, where, $k$ followed by $(b)$ denotes the compact meta-block variant.

\paragraph{Results.}
Table~\ref{tab:cifar} summarizes the results on CIFAR-10 and CIFAR-100 across different backbones and kernel sizes of MCL. Adding MCL on top of the baseline consistently improves performance in all configurations. On CIFAR-10, attaching MCL with $k=3$ already yields noticeable gains. For example, the Top-1 accuracy of ResNet-20 increases from $91.43\%$ to $93.52\%$, while ResNet-110 improves from $93.57\%$ to $94.72\%$. When MCL is combined with stronger data augmentation (baseline + Aug + MCL), the gains are further amplified, achieving up to $2.61\%$ points improvement on ResNet-20 and $2.59\%$ points on ResNet-110.

The improvement is even more pronounced on CIFAR-100. For ResNet-20, incorporating MCL with full augmentation increases Top-1 accuracy from $69.95\%$ to $78.58\%$, corresponding to an $8.63\%$ points gain. WideResNet-28-10 exhibits a similar trend, with accuracy improving from $80.75\%$ to $86.20\%$ under the same setting. Across all backbones, MCL consistently yields strong gains on CIFAR-100, highlighting the benefit of high-order input-adaptive kernels in more challenging multi-class scenarios.

Figure~\ref{fig:c100-accvsparam} provides a comparison of accuracy versus parameter count for different meta-kernel sizes on CIFAR-100. As $k$ increases, MCL progressively improves accuracy while exhibiting a controllable increase in parameter count. This trade-off curve shows that MCL occupies a favorable region, moderate increases in model capacity (e.g., $k=3$) already deliver substantial performance gains.

\subsection{Image Classification on ImageNet}
\label{sec:imagenet}

\begin{table}[t]
\centering
\caption{Performance comparison on the ImageNet dataset using ResNet18 backbone.}
\label{tab:resnet18_in1k}
\resizebox{\linewidth}{!}{
\begin{tabular}{lccc}
\toprule
\textbf{Models} & \textbf{Params} & \textbf{Top-1 (\%)} & \textbf{Top-5 (\%)} \\
\midrule
ResNet18 & 11.69M & 70.25 & 89.38 \\
\midrule
+ SE        & 11.78M & 70.98 ($\uparrow$0.73) & 90.03 ($\uparrow$0.65) \\
+ CBAM      & 11.78M & 71.01 ($\uparrow$0.76) & 89.85 ($\uparrow$0.47) \\
+ ECA       & 11.69M & 70.60 ($\uparrow$0.35) & 89.68 ($\uparrow$0.30) \\
+ CGC       & 11.69M & 71.60 ($\uparrow$1.35) & 90.35 ($\uparrow$0.97) \\
+ WeightNet & 11.93M & 71.56 ($\uparrow$1.31) & 90.38 ($\uparrow$1.00) \\
+ DCD       & 14.70M & 72.33 ($\uparrow$2.08) & 90.65 ($\uparrow$1.27) \\
+ CondConv (8$\times$) & 81.35M & 71.99 ($\uparrow$1.74) & 90.27 ($\uparrow$0.89) \\
+ DY-Conv (4$\times$)   & 45.47M & 72.76 ($\uparrow$2.51) & 90.79 ($\uparrow$1.41) \\
+ ODConv (4$\times$)    & 44.90M & 73.97 ($\uparrow$3.72) & 91.35 ($\uparrow$1.97) \\
+ KW (4$\times$)    & 45.86M & 74.16 ($\uparrow$3.91) & 91.42 ($\uparrow$2.04) \\
\midrule
+ MCL $(k=1)$ & 21.13M & 72.04 ($\uparrow$1.79) & 90.75 ($\uparrow$1.37) \\
+ MCL $(k=3)$ & 36.63M & \textbf{75.55} ($\uparrow$\textbf{5.30}) & \textbf{92.38} ($\uparrow$\textbf{3.00}) \\
+ MCL $(k=5)$ & 158.0M & \textbf{76.86} ($\uparrow$\textbf{6.61}) & \textbf{92.98} ($\uparrow$\textbf{3.60}) \\
\bottomrule
\end{tabular}
}
\end{table}

\begin{table}[t]
    \centering
    \caption{Performance comparison on ImageNet dataset using ResNet50 and ResNet101 backbone network.}
    \label{tab:resnet50-101_in}
    \resizebox{\linewidth}{!}{
    \begin{tabular}{l|ccc|ccc}
    \toprule
        \textbf{Model} & \multicolumn{3}{c|}{\textbf{ResNet50}} & \multicolumn{3}{c}{\textbf{ResNet101}}  \\
        \midrule
         & Params & Top-1 (\%) & Top-5 (\%) & Params & Top-1 (\%) & Top-5 (\%) \\
         \midrule
         Baseline & 25.56M & 76.23 & 93.01 & 44.55M & 77.41 & 93.67\\
         \midrule
         + SE & 28.07M & 77.31 ($\uparrow$1.08) & 93.63 ($\uparrow$0.62) & 49.29M & 78.42 ($\uparrow$1.01) & 94.15 ($\uparrow$0.48) \\
         + CBAM & 28.07M & 77.46 ($\uparrow$1.23) & 93.59 ($\uparrow$0.58) & 49.30M & 78.50 ($\uparrow$1.09) & 94.20 ($\uparrow$0.53) \\
         + ECA & 25.56M & 77.34 ($\uparrow$1.11) & 93.64 ($\uparrow$0.63) & 44.55M & 78.60 ($\uparrow$1.19) & 94.34 ($\uparrow$0.67) \\
         + ODConv & 90.67M & 78.52 ($\uparrow$2.29) & 94.01 ($\uparrow$1.00) & 90.44M & 79.27 ($\uparrow$1.86) & 94.47 ($\uparrow$0.80) \\
         \midrule
         + MCL $(k=1)$ & 177.0M & \textbf{79.25} ($\uparrow$\textbf{3.02}) & \textbf{94.44} ($\uparrow$\textbf{1.43}) & 196.0M & \textbf{80.20} ($\uparrow$\textbf{2.79}) & \textbf{94.77} ($\uparrow$\textbf{1.10}) \\
         + MCL $(k=3)$ & 424.0M & \textbf{79.65} ($\uparrow$\textbf{3.42}) & \textbf{94.46} ($\uparrow$\textbf{1.45}) & 443.0M & \textbf{80.46} ($\uparrow$\textbf{3.05}) & \textbf{94.97} ($\uparrow$\textbf{1.30}) \\
         \bottomrule
         
    \end{tabular}
    }
    
\end{table}

We further evaluate scalability on ImageNet-1K~\cite{imagenet}. We adopt ResNet-18/50/101~\cite{resnet} and compare against representative feature recalibration modules (SE~\cite{se}, CBAM~\cite{cbam}, ECA~\cite{eca}) and dynamic/kernel modulation baselines (CGC~\cite{cgc}, WeightNet~\cite{weightnet}, DCD~\cite{dcd}, CondConv~\cite{condconv}, DY-Conv~\cite{dynamic_conv}, ODConv~\cite{odconv}, KW~\cite{kw}). 
All models are trained for 100 epochs with SGD, using a mini-batch size of 256, momentum $0.9$, and weight decay $5 \times 10^{-4}$. The learning rate is initially set to $0.1$ and decayed to zero with a cosine schedule over the course of training. Training images are resized to $256 \times 256$, followed by random $224 \times 224$ crops and random horizontal flips, as well as per-channel mean–std normalization. At evaluation time, we report single-crop Top-1 and Top-5 accuracy using the $224 \times 224$ center crop. Following previous dynamic convolution studies, we only employ MixUp \cite{cutmix} regularization and label-smoothing for ImageNet, in order to maintain clean and fair performance comparisons.

For MCL, we insert the meta-convolution layer before the fully-connected layer and the backbone network remains unchanged. We consider different meta-kernel sizes $k$ with the polynomial expansion order $n=2$ and we used the compact meta-block design on ImageNet to reduce the parameter cost.

\paragraph{Results.}
Training stability is further evidenced by the standard deviations reported across 3 independent runs in Table~\ref{tab:cifar}, and convergence curves for all evaluated backbones are provided in Appendix~\ref{app:optimization-behavior}. Table~\ref{tab:resnet18_in1k} reports ImageNet results using ResNet-18 \cite{resnet} as the backbone. Standard attention and kernel modulation modules like SE \cite{se}, CBAM \cite{cbam}, ECA \cite{eca}, CGC\cite{cgc}, WeightNet\cite{weightnet}, DCD\cite{dcd} provide moderate gains over the baseline, with improvements typically within $2\%$ of Top-1 accuracy. Dynamic convolution approaches such as CondConv\cite{condconv}, DyConv\cite{dynamic_conv}, ODConv\cite{odconv}, and KW\cite{kw} yield larger improvements, with the best prior configuration (KW (4$\times$)) achieving $74.16\%$ Top-1 accuracy \cite{kw}.

By contrast, MCL delivers substantially larger gains. With $k = 1$, MCL already improves ResNet-18 from $70.25\%$ to $72.04\%$ Top-1 accuracy. Increasing the kernel size to $k=3$ and $k=5$ further boosts performance to $75.55\%$ and $76.86\%$, respectively. The best configuration, MCL with $k=5$, achieves a $6.61\%$ Top-1 improvement and a $3.60\%$ Top-5 improvement over the baseline, surpassing all compared attention and dynamic convolution methods by a clear margin. The MCL configuration with $k{=}3$ reaches $75.55\%$ with $36.63$M parameters, comparable in cost to prior dynamic convolution baselines yet outperforming them by a clear margin. The configuration with $k{=}5$ is included to characterize the upper bound of the accuracy and parameter curve. The gain over a parameter-matched ResNet-50 at $76.23\%$ with $25.6$M parameters confirms that the improvement stems from the polynomial meta-kernel rather than parameter scaling alone. As shown in Figure~\ref{fig:c100-accvsparam}, moderate kernel sizes already occupy a favorable accuracy and parameter region. These results indicate that directly learning a single high-order polynomial meta-kernel is more effective than aggregating multiple static kernels via attention, especially when the input resolution and category diversity are high. A detailed comparison of inference latency, throughput, and peak memory usage against prior dynamic convolution methods is provided in Appendix~\ref{app:latency-memory-overhead}.

Table~\ref{tab:resnet50-101_in} presents the ImageNet results on ResNet-50 and ResNet-101. Similar to the ResNet-18 case, adding SE, CBAM, or ECA to the baseline models yields modest $1\%$ Top-1 gains. ODConv provides larger improvements, achieving $78.52\%$ and $79.27\%$ Top-1 accuracy on ResNet-50 and ResNet-101, respectively. MCL further improves over these strong baselines, with $k = 1$, it reaches $79.25\%$ and $80.20\%$, and with $k = 3$, it achieves $79.65\%$ and $80.46\%$ Top-1 accuracy for ResNet-50 and ResNet-101 respectively. In both backbones, MCL yields around $3\%$ improvement over the vanilla ResNet and about $1\%$ gain over ODConv, confirming that the proposed meta-polynomial kernel generalizes well across deeper architectures.

While MCL $k=3$ yields the largest accuracy gains reported in Table~\ref{tab:resnet50-101_in}, this comes at a substantial increase in parameter count relative to the baseline. We note that the meta-kernel size $k$ is a controllable design choice rather than a fixed cost: MCL$k=1$ already recovers a large fraction of this improvement at a much smaller parameter overhead, as shown for ResNet-18 in Table~\ref{tab:resnet18_in1k}, where MCL$k=1$ reaches 72.04\% Top-1 accuracy with 21.13M parameters, compared to 158.0M parameters for $k=5$. Practitioners operating under tighter parameter budgets can therefore select $k=1$ rather than $k=3$ or $k=5$. This accuracy-parameter trade-off is further characterized in Figure~\ref{fig:c100-accvsparam} and Table~\ref{tab:complexity}. Moreover, as shown in Table~\ref{tab:latency_comparison} (Appendix), the run-time cost of MCL, measured in latency and throughput, scales considerably more gently than its parameter count, since the additional parameters are concentrated in a single meta-network evaluated once per forward pass rather than distributed across repeated operations throughout the backbone.

\subsection{Fine-Grained Visual Classification with MCL}
\label{sec:fgvc}

\begin{table}
    \centering
    \caption{Details of the FGVC datasets used in our experiments.}
    \label{tab:fgvc-dataset}
    \resizebox{\linewidth}{!}{
    \begin{tabular}{lccc}
        \toprule
        \textbf{Dataset} & \textbf{Class} & \textbf{Train} & \textbf{Test} \\
        \midrule
         NABirds \cite{nabirds} & $555$ & $23929$ & $24633$ \\
         CUB-200-2011 \cite{cub} & $200$ & $5994$ & $5794$ \\
         Oxford 102 Flowers \cite{flowers} & $102$ & $2040$ & $6179$ \\
         Oxford-IIIT Pet \cite{pet} & $37$ & $3680$ & $3669$ \\
         \bottomrule
    \end{tabular}
    }
    
\end{table}

Following standard practice for FGVC, all transformer experiments fine-tune ImageNet-pretrained backbones. MCL consistently provides accuracy gains without introducing additional training instability, regardless of the host architecture. In contrast to Transformers, MCL demonstrates particular effectiveness on smaller datasets. We evaluate transfer learning on NABirds~\cite{nabirds}, CUB-200-2011~\cite{cub}, Oxford-102 Flowers~\cite{flowers}, and Oxford-IIIT Pets~\cite{pet}. These datasets are widely adopted for FGVC evaluation with pretrained networks precisely because of their smaller scale, and they provide a useful testbed for demonstrating MCL's effectiveness in data-constrained scenarios where large architectures such as Vision Transformers struggle to train from scratch. We fine-tune ImageNet-pretrained Swin-B~\cite{swin} and ViT-B/16~\cite{vit} with and without MCL, inserting a single module before the classifier. 

Since Swin-B keeps its internal feature maps in a spatial, channel-last format like $H \times W \times C$ rather than a flattened token sequence, MCL's convolutional operators are applied after permuting this tensor to a standard channel-first layout, and the output is permuted back before the classification head. For ViT-B/16, the backbone instead outputs a flat sequence of patch tokens rather than a spatial feature map. To apply MCL, which requires convolutional operators over a spatial input, the patch token sequence is reshaped into its corresponding spatial grid before being passed through MCL, and the resulting output is flattened back into a token sequence before the classification head. This is analogous to the permutation used for Swin-B, but requires reshaping from a flat sequence rather than permuting an already-spatial tensor. 

We follow the commonly adopted fine-tuning schedule for FGVC, using $5$ epochs of warm-up followed by cosine-annealed learning rates. The initial learning rate is set to $5 \times 10^{-4}$ for all the datasets along with a SGD optimizer using momentum of $0.9$ and a weight decay of $1 \times 10^{-4}$. The input image is first scaled up to the size of $510\times510$ for Swin Transformer and $550\times550$ for ViT. After that random cropping is applied to attain fixed-sized image of size $384\times384$ for Swin Transformer and $448\times448$ for ViT during training. Center crop is used at the test time to obtain the same sized images. Random flipping and random Gaussian blur are applied during the training process for all the datasets. All the models are trained for $50$ epochs with a batch size of $16$ for all the datasets presented in Table \ref{tab:fgvc-dataset}.
Unless stated otherwise, we use $k=1$ and $n=2$.

\begin{table}
    \centering
    \caption{Performance comparison of the proposed MCL on FGVC task using commonly used FGVC datasets.}
    \label{tab:fgvc_results}
    \resizebox{0.9\columnwidth}{!}{
    \begin{tabular}{c|llc}
        \toprule
        \textbf{Dataset} & \textbf{Method} & \textbf{Backbone} & \textbf{Accuracy (\%)}  \\
        \midrule
        \multirow{5}{*}{\textbf{NABirds}} & IELT & ViT-B/16 & $90.8$ \\
        & MP-FGVC & ViT-B/16      & $91.0$ \\
        & ACC-ViT & ViT-B/16      & $91.4$ \\
        & MPSA & Swin-B        & $92.5$ \\\cmidrule{2-4}
        & \textbf{MCL} $(k=1)$ & Swin-B        & $\mathbf{92.8}$ \\
        \midrule
        \multirow{5}{*}{\textbf{CUB-200-2011}} & MP-FGVC  & ViT-B/16      & $91.8$ \\
        & ACC-ViT  & ViT-B/16      & $91.8$ \\
        & DACL & ResNet50-ViT  & $92.0$ \\
        & MPSA  & Swin-B        & $\mathbf{92.8}$ \\\cmidrule{2-4}
        & \textbf{MCL} $(k=1)$ & Swin-B        & $92.7$ \\
        \midrule
        \multirow{5}{*}{\textbf{Oxford 102 Flowers}} & InterAct & DenseNet-161  & $96.4$\\
        & ViT & ViT-B/16      & $99.4$ \\
        & CvT & CvT-21        & $99.5$ \\
        & IELT & ViT-B/16      & $99.6$ \\\cmidrule{2-4}
        & \textbf{MCL} $(k=1)$ & Swin-B & $\mathbf{99.6}$ \\
        \midrule
        \multirow{5}{*}{\textbf{Oxford-IIIT Pet}} & InterAct & DenseNet-161  & $ 93.5$ \\
        & ViT & ViT-B/16      & $93.8$ \\
        & CvT & CvT-21        & $94.7$ \\
        & IELT & ViT-B/16     & $95.2$ \\\cmidrule{2-4}
        & \textbf{MCL} $(k=1)$ & Swin-B & $\mathbf{95.9}$ \\
        \bottomrule
    \end{tabular}
    }
    
\end{table}

\begin{table}[t]
\centering
\caption{Performance of the proposed MCL across various FGVC benchmarks using different transformer-based architectures. }
    \label{tab:backbone_improvement}

    \resizebox{0.9\columnwidth}{!}{
    \begin{tabular}{c|c|cccc}
        \toprule
        \textbf{Backbone} & \textbf{Method} & \textbf{CUB} & \textbf{NABirds} & \textbf{Flowers} & \textbf{Pet}\\
        \midrule
        \multirow{3}{*}{Swin-B}
          & Baseline    & $91.93$ & $91.99$ & $99.41$ & $95.09$ \\
          & MCL $(k=1)$         & $92.72$ & $92.83$ & $99.64$ & $95.91$ \\
          & Improvement & $0.79$ & $0.84$ & $0.24$ & $0.82$ \\ \midrule
        \multirow{3}{*}{ViT-B/16}
          & Baseline    & $88.90$ & $87.55$ & $97.06$ & $94.44$ \\
          & MCL $(k=1)$         & $90.35$ & $89.38$ & $99.17$ & $94.55$ \\
          & Improvement & $1.45$ & $1.83$ & $2.11$ & $0.11$ \\ 
          \bottomrule
    \end{tabular}
    }
    
\end{table}

\paragraph{Results on FGVC.}
Tables~\ref{tab:fgvc_results} and~\ref{tab:backbone_improvement} summarize the performance of MCL on NABirds, CUB-200-2011, Oxford-102 Flowers and Oxford-IIIT Pets. When combined with Swin-B, MCL consistently improves over the strong Swin-B baselines on all four datasets. For instance, the Top-1 accuracy on CUB-200-2011 increases from $91.93\%$ to $92.72\%$, and on NABirds from $91.99\%$ to $92.83\%$, while on Oxford-102 Flowers and Oxford-IIIT Pets, MCL achieves $99.64\%$ and $95.91\%$, respectively. Moreover, the proposed method demonstrates comparative or improved performance over state-of-the-art methods like IELT\cite{ielt}, MP-FGVC\cite{mp-fgvc}, MPSA\cite{mpsa}, ACC-ViT\cite{acc-vit}, DACL\cite{dacl}, InterAct\cite{interact} and CvT\cite{cvt}. These gains, although modest in absolute terms due to the high baseline performance, demonstrate that MCL remains effective even in regimes where the backbone already captures rich fine-grained cues.

On ViT-B/16, the improvements are even more pronounced. Across the four FGVC datasets, MCL yields gains of up to $2.11\%$ on Oxford-102 Flowers over the baseline ViT-B/16. The average improvement across all datasets is substantial for both Swin-B and ViT-B/16, confirming that the meta-polynomial kernel can be seamlessly integrated into transformer-based architectures and provides consistent benefits in fine-grained recognition tasks.

\begin{figure}
    \centering
    \includegraphics[width=0.9\linewidth]{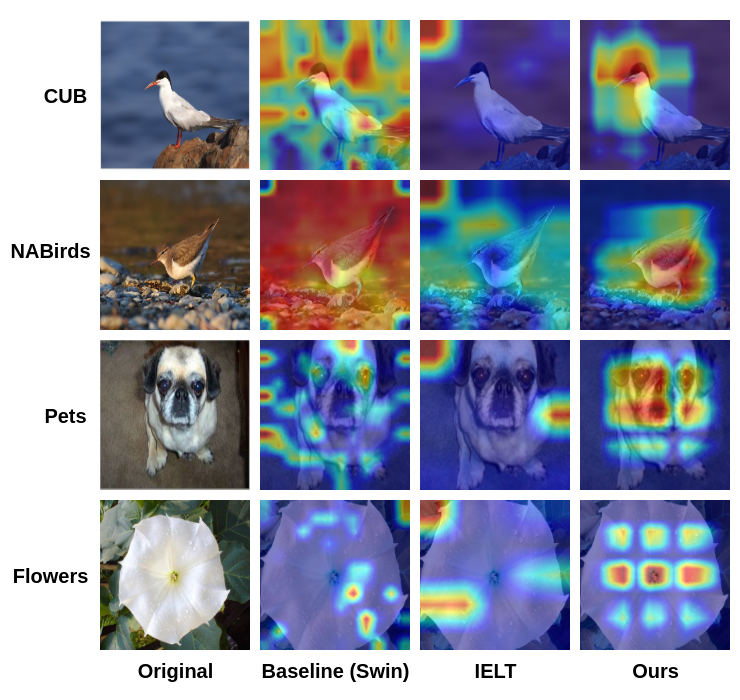}
    \caption{Visual comparison of the proposed method using Grad-Cam visualization.}
    \label{fig:gradcam}
\end{figure}

\paragraph{Visual comparison.}
To better understand the qualitative effect of MCL, we visualize class-discriminative regions using Grad-CAM\cite{gradcam} on representative samples from the four FGVC datasets. Although MCL and competing methods achieve similar quantitative performance on some datasets, the spatial activation patterns differ noticeably, suggesting that the polynomial meta-kernel produces qualitatively distinct feature responses even when top-1 accuracy margins are small. As illustrated in Figure~\ref{fig:gradcam}, models equipped with MCL attend more tightly to object regions and capture semantically meaningful parts, while baselines without MCL often exhibit more diffuse or background-focused activations. For example, in bird and flower categories, MCL encourages the network to focus on characteristic patterns such as heads, wings, or petals rather than surrounding background, which aligns with the observed improvements in classification accuracy. These visualizations suggest that the polynomial meta-kernel enables more structured and task-relevant feature responses.

\begin{table}
    \centering
    \caption{Comparison of computational cost of the proposed method while using transformer architectures.}
    \label{tab:complexity}
    \resizebox{0.8\linewidth}{!}{
    \begin{tabular}{lcccc}
        \toprule
        \multirow{2}{*}{\textbf{Method}} & \multirow{2}{*}{\textbf{Backbone}} & \textbf{Input} & \textbf{Param} & \textbf{FLOPs}  \\
        & & \textbf{Size} & \textbf{(M)} & \textbf{(G)} \\
        \midrule
        ViT & ViT-B/16 & $448$ & $86.4$ & $78.5$ \\
        IELT & ViT-B/16 & $448$ & $93.5$ & $73.2$ \\
        TransFG  & ViT-B/16 & $448$ & $86.4$ & $130.2$ \\
        RAMS-Trans  & ViT-B/16 & $448$ & $86.4$ & $157.4$ \\
        ACC-ViT  & ViT-B/16 & $448$ & $87.0$ & $162.9$ \\
        \midrule
        Swin  & Swin-B & $384$ & $87.9$ & $47.2$ \\
        MPSA & Swin-B & $384$ & $97.4$ & $49.2$ \\
        ViT-Net  & Swin-B & $448$ & $92.2$ & $65.6$ \\
        \midrule
        \textbf{MCL} $(k=1)$ & Swin-B & $384$ & $127.0$ & $52.9$ \\
        \textbf{MCL} $(k=1)$ & ViT-B/16 & $448$ & $109.0$ & $84.8$ \\
        \bottomrule
        
    \end{tabular}
    }
    
\end{table}

\paragraph{Computational cost.}
We further quantify the efficiency of MCL in the FGVC setting by comparing parameter counts and FLOPs against representative transformer-based FGVC methods. The results are presented in Table~\ref{tab:complexity}. On ViT-B/16 with $448\times448$ inputs, MCL increases the baseline from $86.4$M/$78.5$G to $109.0$M/$84.8$G (Params/FLOPs). On Swin-B with $384\times384$ inputs, MCL raises the baseline from $87.9$M/$47.2$G to $127.0$M/$52.9$G. While MCL introduces additional parameters due to the meta-kernel parameterization, the FLOPs overhead remains moderate relative to the underlying backbone.

More importantly, Table~\ref{tab:complexity} indicates that MCL achieves favorable compute efficiency when compared to several FGVC-specific transformer pipelines. In particular, the FLOPs of MCL-equipped Swin-B ($52.9$G) remain close to the base Swin-B ($47.2$G) and are substantially lower than alternatives that rely on heavier architectural modifications or multi-stage processing like, TransFG\cite{transfg} and RAMS-Trans\cite{rams-trans}. Similarly, for ViT-B/16, MCL incurs only a small FLOPs increase over the baseline while avoiding the significantly higher compute budgets reported by some prior FGVC methods. These results suggest that MCL provides a practical accuracy--efficiency trade-off, delivering consistent gains in fine-grained recognition without resorting to computationally expensive FGVC-specific designs.

\section{Conclusion}
\label{sec:conclusion}
We introduced the MCL, a dynamic convolution mechanism that generates an input-conditioned convolution kernel via a structured high-order polynomial expansion. Unlike prior dynamic convolution approaches that rely on linear mixtures of multiple basis kernels, MCL produces a single adaptive kernel without kernel-mixture attention, eliminating the need for temperature scheduling and alleviating the difficulty of jointly optimizing multiple basis kernels and their input-dependent mixing weights as the mixture size grows. 

Experiments on ImageNet-1K and CIFAR demonstrate consistent improvements over vanilla networks and competing methods, and additional results on fine-grained datasets show that the approach generalizes to transformer backbones as well. Overall, MCL offers a practical and scalable alternative for input-adaptive parameterization, and future work may further reduce overhead, explore deeper placement strategies across network stages, and extend evaluation to broader downstream tasks such as object detection and semantic segmentation. In this work, MCL is added only at the end of the backbone, which fits classification since the module only needs to refine one final representation. Detection and segmentation instead need features at many spatial resolutions, so MCL could instead be added at multiple stages of the backbone so it can generate kernels at different resolutions. We have not tested this multi-stage setting and leave it for future work. Additionally, while larger kernel sizes offer favorable accuracy gains, they also introduce non-trivial parameter overhead. Reducing this cost through parameter-efficient techniques such as tensor factorization could further improve the scalability of the method.

\section*{Acknowledgment}
This study was supported by research fund from Chosun University, 2026.

\bibliographystyle{IEEEtran}
\bibliography{bibliography}


\clearpage


\setcounter{section}{0}
\setcounter{subsection}{0}
\renewcommand{\thesection}{\Alph{section}}
\renewcommand{\thesubsection}{\thesection.\arabic{subsection}}

\refstepcounter{section}
\section*{\thesection. Ablation Study on meta-kernel size $k$ and order of expansion $n$}
\label{sec:appendix-k-n}

This appendix analyzes the sensitivity of MCL to two key design factors: the meta-kernel size $k$ and the polynomial expansion order $n$. Tables~\ref{tab:k-n} and~\ref{tab:k-n-b} report Top-1 accuracy and parameter counts on CIFAR-100 using a ResNet-20 backbone under two meta-block realizations.

\refstepcounter{subsection}
\subsection*{\thesubsection. Basic Meta-Block}
\label{app:basic-meta-block}

Table~\ref{tab:k-n} shows that increasing the expansion order $n$ generally improves performance for moderate kernels. In particular, for $k\in\{1,3\}$, accuracy increases monotonically as $n$ grows, indicating that higher-order interactions contribute positively to discriminative capacity. For $k=3$, the gains are consistent yet modest, suggesting that a moderate spatial receptive field is already expressive and benefits primarily from additional higher-order modulation.

In contrast, enlarging the kernel to $k=5$ under the basic meta-block substantially increases the parameter count while providing limited additional accuracy. This behavior suggests that, without further structural constraints, scaling $k$ can quickly become an inefficient way to improve representation. Overall, the basic meta-block favors moderate kernel sizes, e.g., $k=3$, with small-to-medium expansion orders.

\begin{table}[h]
    \centering
    \caption{Classification accuracy of ResNet-20 backbone network using MCL with varying meta-kernel size $k$ and order of polynomial expansion $2^n$ on the CIFAR-100 dataset. Here, MCL is constructed with only the basic meta-block.}
    \label{tab:k-n}
    \begin{tabular}{c|cc|cc|cc}
        \toprule
        & \multicolumn{2}{c|}{$k=1$} 
        & \multicolumn{2}{c|}{$k=3$} 
        & \multicolumn{2}{c}{$k=5$} \\
        \hline
        & \#param & Top-1 
        & \#param & Top-1 
        & \#param & Top-1 \\
        \midrule
        $n=1$ & $0.35$M & $71.94$ & $3.63$M & $77.69$ & $24.35$M & $77.41$ \\
        $n=2$ & $0.42$M & $72.86$ & $3.70$M & $77.88$ & $24.42$M & $77.92$ \\
        $n=3$ & $0.49$M & $\mathbf{73.66}$ & $3.78$M & $\mathbf{78.19}$ & $24.49$M & $\mathbf{77.94}$ \\
        \bottomrule
    \end{tabular}
\end{table}

\refstepcounter{subsection}
\subsection*{\thesubsection. Compact Meta-Block}
\label{app:compact-meta-block}

Table~\ref{tab:k-n-b} demonstrates that the Compact Meta-Block markedly improves parameter efficiency, especially when using larger kernels. Compared with the Basic Meta-Block design, the Compact Meta-Block substantially reduces the cost of spatial modeling, allowing $k=5$ to become viable without incurring prohibitive parameters. Notably, the best-performing Compact Meta-Block configuration is achieved at a moderate expansion order, typically $n=2$, while further increasing $n$ does not necessarily yield additional gains and may even degrade accuracy for certain kernel sizes. This observation suggests that, under the Compact Meta-Block constraints, excessive expansion order can introduce optimization difficulty or reduce effective capacity due to over-constrained intermediate representations.

\begin{table}[h]
    \centering
    \caption{Classification accuracy of ResNet-20 backbone network using MCL with varying meta-kernel size $k$ and order of polynomial expansion $2^n$ on the CIFAR-100 dataset. Here, MCL is constructed with only the compact meta-block.}
    \label{tab:k-n-b}
    \begin{tabular}{c|cc|cc|cc}
        \toprule
        & \multicolumn{2}{c|}{$k=1$} 
        & \multicolumn{2}{c|}{$k=3$} 
        & \multicolumn{2}{c}{$k=5$} \\
        \hline
        & \#param & Top-1 
        & \#param & Top-1 
        & \#param & Top-1 \\
        \midrule
        $n=1$ & $0.28$M & $\mathbf{70.93}$ & $0.59$M & $76.13$ & $2.49$M & $77.83$ \\
        $n=2$ & $0.28$M & $70.38$ & $0.60$M & $\mathbf{76.57}$ & $2.49$M & $\mathbf{78.31}$ \\
        $n=3$ & $0.29$M & $70.64$ & $0.60$M & $75.86$ & $2.50$M & $78.07$ \\
        \bottomrule
    \end{tabular}
\end{table}

\refstepcounter{subsection}
\subsection*{\thesubsection. Latency and Memory Overhead Comparison}
\label{app:latency-memory-overhead}

Table~\ref{tab:latency_comparison} presents an inference-time comparison of the proposed MCL against the baseline backbone and representative dynamic convolution baselines, namely KW$(4\times)$~\cite{kw} and ODConv$(4\times)$~\cite{odconv}, on the ImageNet test set. The reported latency, throughput, and peak memory metrics are obtained using a batch size of $128$, and each value is computed over $50$ batches to provide a stable estimate of runtime behavior. As indicated in the table, lower latency and peak memory are preferred, whereas higher throughput is desirable.

For ResNet18, the baseline model achieves $5.23 \pm 1.56$ ms latency, $24472.18$ images/sec throughput, and $910.82$ MB peak memory. Relative to this reference, MCL introduces only a moderate inference overhead: MCL$(k=1)$ attains $7.68 \pm 1.86$ ms and $16659.78$ images/sec with $992.38$ MB peak memory, while MCL$(k=3)$ yields $6.75 \pm 2.13$ ms and $18957.37$ images/sec with $1909.15$ MB peak memory. In contrast, KW$(4\times)$ and ODConv$(4\times)$ incur substantially larger runtime costs, reaching $91.67 \pm 0.56$ ms and $35.04 \pm 1.71$ ms latency, respectively, with markedly lower throughput. These results show that, on a lightweight backbone, MCL remains much closer to the efficiency profile of the static baseline than prior kernel-aggregation methods.

A similar trend is observed on ResNet50. The baseline records $10.24 \pm 1.82$ ms latency, $12503.45$ images/sec throughput, and $1622.69$ MB peak memory. MCL$(k=1)$ and MCL$(k=3)$ increase latency only modestly to $11.73 \pm 2.12$ ms and $11.54 \pm 1.59$ ms, respectively, while sustaining high throughput of $10908.65$ and $11087.52$ images/sec. By comparison, KW$(4\times)$ and ODConv$(4\times)$ exhibit much heavier computational overhead, with latencies of $280.43 \pm 1.73$ ms and $125.32 \pm 0.24$ ms and correspondingly low throughput of $456.44$ and $1021.37$ images/sec.

Overall, MCL delivers input-adaptive convolution with a considerably smaller runtime penalty than methods that rely on aggregating multiple kernels through attention. This observation is consistent with the core design of MCL, which directly generates a single input-conditioned meta-kernel rather than performing expensive multi-kernel composition.

\begin{table}[t]
\centering
\small
\setlength{\tabcolsep}{4pt}
\renewcommand{\arraystretch}{1.15}
\caption{Latency and memory overhead comparison of the proposed method on ImageNet test set. $\downarrow$ indicates lower is better and $\uparrow$ indicates higher is better.}
\label{tab:latency_comparison}
\begin{tabular}{l|c|c|c}
\toprule
\textbf{Model}
& \makecell{\textbf{Latency (ms)} \\ $\downarrow$}
& \makecell{\textbf{Throughput} \\ \textbf{(images/sec)} \\ $\uparrow$}
& \makecell{\textbf{Peak Mem} \\ \textbf{(MB)} \\ $\downarrow$} \\
\midrule
ResNet18            & $5.23  \pm 1.56$ & $24472.18$ & $910.82$  \\
\midrule
+ KW$(4\times)$     & $91.67 \pm 0.56$ & $1396.30$  & $2736.41$ \\
+ ODConv$(4\times)$ & $35.04 \pm 1.71$ & $3653.18$  & $6376.81$ \\
\midrule
+ MCL$(k=1)$        & $7.68  \pm 1.86$ & $16659.78$ & $992.38$  \\
+ MCL$(k=3)$        & $6.75  \pm 2.13$ & $18957.37$ & $1909.15$ \\
\midrule
ResNet50            & $10.24 \pm 1.82$ & $12503.45$ & $1622.69$ \\
\midrule
+ KW$(4\times)$     & $280.43\pm 1.73$ & $456.44$   & $3707.65$ \\
+ ODConv$(4\times)$ & $125.32\pm 0.24$ & $1021.37$  & $7547.10$ \\
\midrule
+ MCL$(k=1)$        & $11.73 \pm 2.12$ & $10908.65$ & $3293.92$ \\
+ MCL$(k=3)$        & $11.54 \pm 1.59$ & $11087.52$ & $22196.76$ \\
\bottomrule
\end{tabular}
\end{table}

\refstepcounter{subsection}
\subsection*{\thesubsection. Optimization Behavior and Run-to-Run Stability}
\label{app:optimization-behavior}

\begin{figure*}[t]
    \centering
    \includegraphics[width=\textwidth]{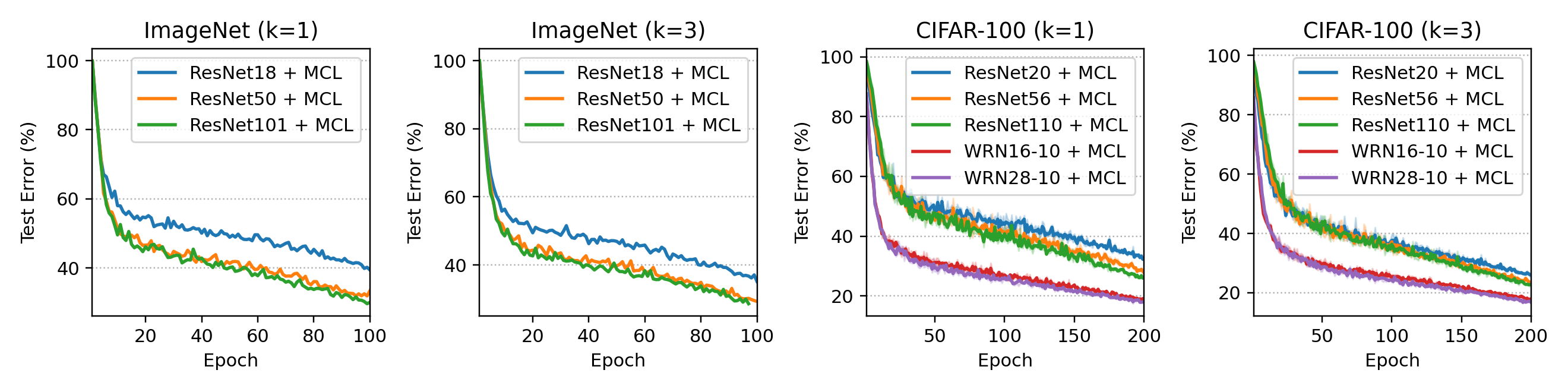}
    \caption{Convergence behavior of MCL on ImageNet and CIFAR-100 under different meta-kernel sizes. The first two panels show the test-error trajectories of ResNet18, ResNet50, and ResNet101 equipped with MCL on ImageNet for $k=1$ and $k=3$, respectively. The last two panels show the corresponding CIFAR-100 results for ResNet20, ResNet56, ResNet110, WRN16-10, and WRN28-10. For CIFAR-100, solid lines denote the mean of three independent runs and the shaded bands indicate the standard deviation. Across datasets, architectures, and kernel settings, MCL exhibits smooth convergence and low run-to-run variability, supporting the claim that the proposed layer is stable and easy to optimize in practice.}
    \label{fig:mcl_convergence}
\end{figure*}

In addition to the ablations on meta-kernel size $k$ and expansion order $n$, we further examine the optimization behavior of MCL to directly address the claim made in the main text that the proposed formulation is easier to optimize than prior attention-based dynamic convolution methods. The convergence plots in Fig.~\ref{fig:mcl_convergence} provide empirical evidence of a stable training procedure.

First, across ImageNet, MCL exhibits stable and well-behaved convergence for all evaluated backbones. For both $k=1$ and $k=3$, the test error decreases rapidly during the early stage of training and continues to improve smoothly over later epochs for ResNet18, ResNet50, and ResNet101. Importantly, we do not observe unstable oscillations or optimization collapse as network depth increases, which suggests that the proposed meta-polynomial kernel generation remains trainable even when integrated into larger backbone architectures. This observation is consistent with the design rationale of MCL in Sections~\ref{subsec:meta-kernel} and~\ref{subsec:implementation}, where the adaptive kernel is learned end-to-end through a single meta-network without requiring auxiliary optimization heuristics such as temperature annealing.

Second, the CIFAR-100 results indicate that this optimization behavior is also stable across independent runs. In Fig.~\ref{fig:mcl_convergence}, the solid curves for CIFAR-100 denote the mean test error over three independent runs, and the shaded regions represent the corresponding standard deviation. Across ResNet20, ResNet56, ResNet110, WRN16-10, and WRN28-10, the variance bands remain relatively narrow throughout training for both $k=1$ and $k=3$. This low run-to-run variability suggests that MCL is not overly sensitive to random initialization or stochasticity in optimization, and that the observed performance gains are reproducible rather than arising from isolated favorable runs.

\refstepcounter{subsection}
\subsection*{\thesubsection. Discussion}
\label{app:discussion}

Taken together, Tables~\ref{tab:k-n}--\ref{tab:k-n-b} highlight a clear trade-off. Increasing $n$ is beneficial up to a small value, after which returns diminish and may become unstable depending on the meta-block structure. Increasing $k$ can improve accuracy, but is only practical when paired with a parameter-efficient realization such as the Compact Meta-Block. In practice, a strong accuracy--efficiency balance is obtained with $k=3$ and a small expansion order, e.g., $n\in\{2,3\}$, for the Basic Meta-Block design, while the Compact Meta-Block enables larger kernels, e.g., $k=5$, to achieve competitive accuracy with substantially reduced parameter overhead.

\end{document}